\documentclass[letterpaper]{article} 
\usepackage{aaai2026}  
\usepackage{times}  
\usepackage{helvet}  
\usepackage{courier}  
\usepackage[hyphens]{url}  
\usepackage{graphicx} 
\usepackage[numbers,square,comma,sort&compress]{natbib} 
\usepackage{caption} 
\usepackage{xcolor}
\usepackage{booktabs}
\usepackage{multirow}
\usepackage{amsmath}
\usepackage{amssymb}
\usepackage{algorithm}
\usepackage{algpseudocode}
\usepackage{makecell}

\usepackage{newfloat}
\usepackage{listings}
\DeclareCaptionStyle{ruled}{labelfont=normalfont,labelsep=colon,strut=off} 
\floatstyle{ruled}
\newfloat{listing}{tb}{lst}{}
\floatname{listing}{Listing}
\title{ReTrack: Evidence-Driven Dual-Stream Directional Anchor Calibration Network for Composed Video Retrieval}
\author{
    Zixu Li\textsuperscript{\rm 1},
    Yupeng Hu\textsuperscript{\rm 1}\thanks{Corresponding author.},
    Zhiwei Chen\textsuperscript{\rm 1},
    Qinlei Huang\textsuperscript{\rm 1}, 
    Guozhi Qiu\textsuperscript{\rm 1},
    Zhiheng Fu\textsuperscript{\rm 1},
    Meng Liu\textsuperscript{\rm 2}
}
\affiliations{
    \textsuperscript{\rm 1}School of Software, Shandong University\\ \textsuperscript{\rm 2}School of Computer Science and Technology, Shandong Jianzhu University\\
    \{lizixu.cs, zivczw, fuzhiheng8, mengliu.sdu\}@gmail.com, 
    \{hql, qiugz\}@mail.sdu.edu.cn, huyupeng@sdu.edu.cn,
}

\usepackage{bibentry}

\begin{document}
\maketitle

\begin{abstract}
With the rapid growth of video data, Composed Video Retrieval (CVR) has emerged as a novel paradigm in video retrieval and is receiving increasing attention from researchers. Unlike unimodal video retrieval methods, the CVR task takes a multi-modal query consisting of a reference video and a piece of modification text as input. The modification text conveys the user's intended alterations to the reference video. Based on this input, the model aims to retrieve the most relevant target video. In the CVR task, there exists a substantial discrepancy in information density between video and text modalities. Traditional composition methods tend to bias the composed feature toward the reference video, which leads to suboptimal retrieval performance. This limitation is significant due to the presence of three core challenges: (1) \textbf{modal contribution entanglement}, (2) \textbf{explicit optimization of composed features}, and (3) \textbf{retrieval uncertainty}. To address these challenges, we propose the evidence-d\textbf{R}iv\textbf{E}n dual-s\textbf{T}ream di\textbf{R}ection\textbf{A}l an\textbf{C}hor calibration networ\textbf{K} (\textbf{ReTrack}). ReTrack is the first CVR framework that improves multi-modal query understanding by calibrating directional bias in composed features. It consists of three key modules: \textit{Semantic Contribution Disentanglement}, \textit{Composition Geometry Calibration}, and \textit{Reliable Evidence-driven Alignment}. Specifically, ReTrack estimates the semantic contribution of each modality to calibrate the directional bias of the composed feature. It then uses the calibrated directional anchors to compute bidirectional evidence that drives reliable composed-to-target similarity estimation. Moreover, ReTrack exhibits strong generalization to the Composed Image Retrieval (CIR) task, achieving SOTA performance across three benchmark datasets in both CVR and CIR scenarios. Codes are available at https://github.com/Lee-zixu/ReTrack
\end{abstract}

\begin{figure}[t]
\begin{center}
\includegraphics[width=\linewidth]{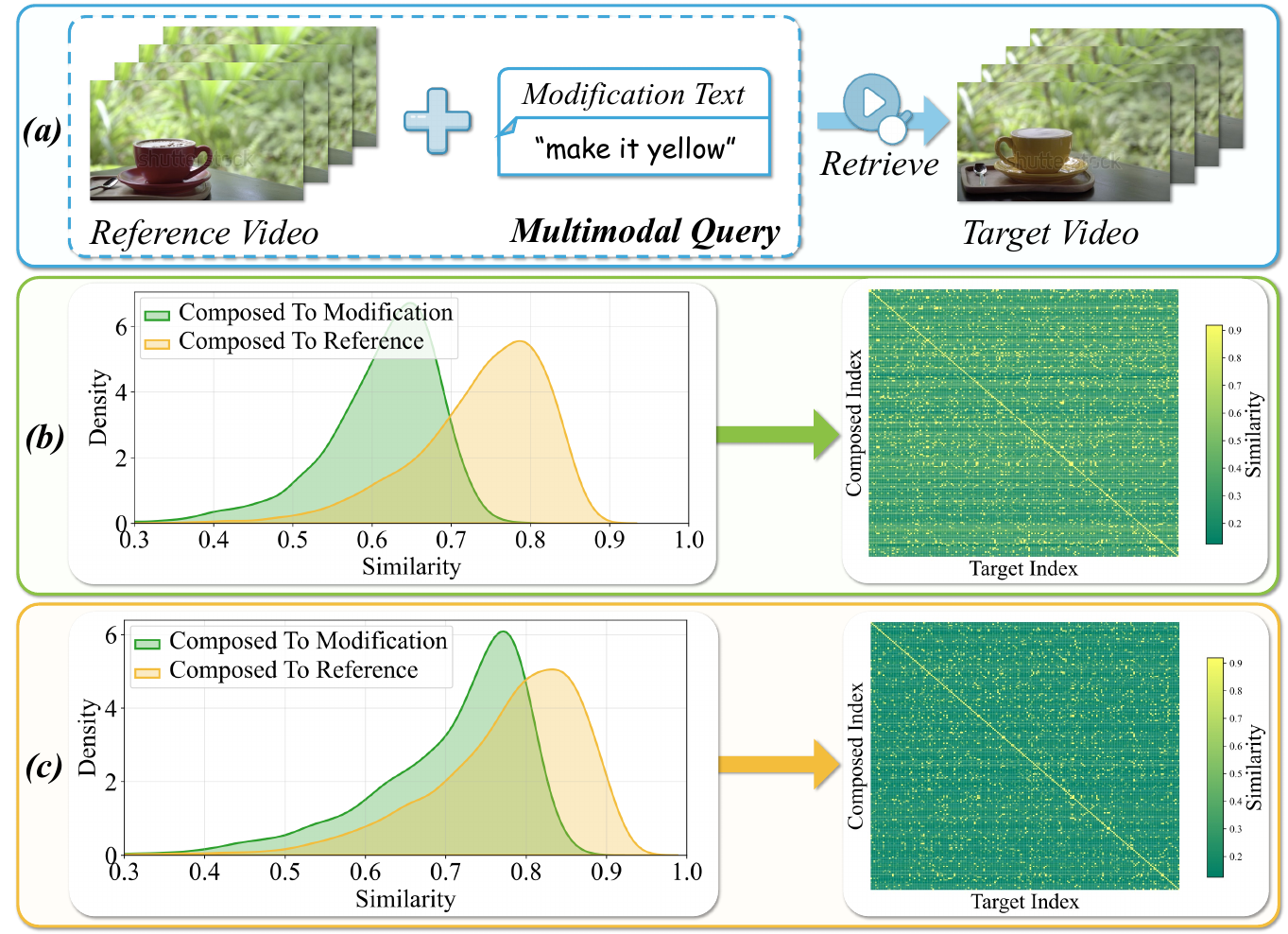}
\end{center}
   \caption{(a) illustrates a typical CVR example. (b) highlights the directional bias issue in existing methods, where the similarity between the composed feature and the target video becomes indistinguishable from that of certain negative candidates, degrading retrieval performance. (c) demonstrates that our method effectively mitigates directional bias, producing a clear separation between the composed feature's similarity to the target and all negative samples.}
\label{fig:intro}
\end{figure}
\section{Introduction}\label{sec:intro}
With the rapid expansion of video data~\cite{wang2024twin}, video retrieval has become a central research focus in the field of multimodal processing~\cite{lu2025does, bi2025llava,zhou2025dragflow,xu2025noisy}, information retrieval~\cite{cheng2026enhancing,xie2025chat}, and multimedia learning~\cite{song3,chen2025autoneural,xie2026conquer,song2}. To meet growing demands for flexible queries, Ventura et al.\cite{covr} proposed Composed Video Retrieval (CVR), which has since gained significant attention\cite{covr-2, covr_enrich, fdca,REFINE}. As shown in Figure~\ref{fig:intro}(a), unlike traditional unimodal video retrieval~\cite{xie2026hvd,xie2026delving}, CVR retrieves the most relevant target video from a large-scale database using a multi-modal query comprising a reference video and a modification text. As a fundamental task in multi-modal interaction, CVR supports real-world applications such as multimodal reasoning~\cite{meng2026tri,Feng2024CLIPCleaner,sun2024robust,yu2025vismem,bi2025prismselfpruningintrinsicselection,ni2025recondreamrRL,ni2025recondreamer,ERASE}, and intelligent interaction system~\cite{yu2025iidm,li2023hong,jiang2025stg,jiang2025transforming,yu2025visualizing,lu2024robust,bi2025cot,wen2023syreanet,STABLE}.

However, due to the overlooked directional bias in the composed feature, CVR remains at an early stage. Specifically, the video modality typically captures rich temporal and visual information, while the text modality conveys semantics concisely, resulting in a notable discrepancy in information density. Therefore, existing CVR methods~\cite{covr, covr-2, covr_enrich} that utilize unified encoders (e.g., BLIP, BLIP-2) to encode video and text data tend to exhibit semantic bias. As shown in Figure~\ref{fig:intro}(b), the composed features generated by existing methods often exhibit excessively high similarity to the reference video \textbf{(yellow area)} while showing low similarity to the modification text \textbf{(green area)}. As illustrated in the similarity matrix on the right, this leads to the composed feature having a similarity to the positive target video that is close to that of certain negative candidates, ultimately resulting in degraded retrieval accuracy.

To address the directional bias in the composed feature, we propose a strategy based on a dual-stream directional anchor to explicitly calibrate the composed feature, enabling accurate integration of cross-modal semantics. As illustrated in Figure~\ref{fig:intro}(c), the composed feature generated by our method exhibits comparable similarity to both the reference and modification semantics, and achieves improved discriminability among candidate target videos. However, implementing this strategy involves three primary challenges.
\textbf{(1) Modal contribution entanglement.} Correcting directional bias requires identifying the semantic contributions of each modality. Nevertheless, due to the entangled nature of these semantics and the lack of explicit supervision, disentangling the semantic contributions from different modalities within the composed feature constitutes the first challenge.
\textbf{(2) Explicit optimization of composed features.} Once the semantic contributions have been identified, the second challenge lies in evaluating whether the composed feature exhibits semantic directional bias based on the current semantic contributions, and performing explicit calibration accordingly.
\textbf{(3) Retrieval uncertainty.} Similar to Composed Image Retrieval (CIR), the CVR task also relies on triplet data, which is expensive to annotate and often contains a large number of visually or semantically similar candidate videos~\cite{fdca}. Such videos introduce substantial uncertainty in retrieving the correct target video. Consequently, relying solely on the similarity between the composed feature and candidate videos may be insufficient for accurate retrieval. The third challenge, therefore, is how to evaluate the reliability of similarity estimation to achieve precise retrieval.

To address the above challenges, we propose the evidence-d\textbf{R}iv\textbf{E}n dual-s\textbf{T}ream di\textbf{R}ection\textbf{A}l an\textbf{C}hor calibration networ\textbf{K} (\textbf{ReTrack}), which calibrates the directional bias of the composed feature and leverages calibrated directional anchors to compute bidirectional evidence for reliable composed-to-target similarity estimation. Specifically, (1) to resolve the issue of \textbf{modal contribution entanglement}, we introduce \textit{Semantic Contribution Disentanglement}, which separates visual and textual semantic contributions within the composed feature to support subsequent bias correction; (2) to address \textbf{explicit optimization of composed features} challenge, we propose \textit{Composition Geometry Calibration}, which builds directional anchors based on modality semantic contribution and reconstructs the composed feature to eliminate directional bias; (3) to mitigate \textbf{retrieval uncertainty}, we design \textit{Reliable Evidence-driven Alignment}, which derives bidirectional evidence from interactions between anchors and target features, enabling adaptive weighting of high-credible samples and robust alignment between composed and target features.

In summary, our contributions include:
\begin{itemize}
	\item We propose a novel Composed Video Retrieval (CVR) framework named ReTrack. To the best of our knowledge, it is the first CVR model that improves multi-modal query understanding by correcting the directional bias in the composed feature.
	\item ReTrack enables secondary construction of the composed feature by disentangling the semantic contribution, allowing for fine-grained adjustment of its spatial position and directional bias. It further performs similarity reliability estimation through evidence learning to achieve precise composed feature optimization.
	\item Extensive experiments conducted on three widely-used benchmark datasets, covering both CVR and CIR tasks, demonstrate the superiority of our proposed ReTrack.
\end{itemize}

\begin{figure*}
\begin{center}
\includegraphics[width=\linewidth]{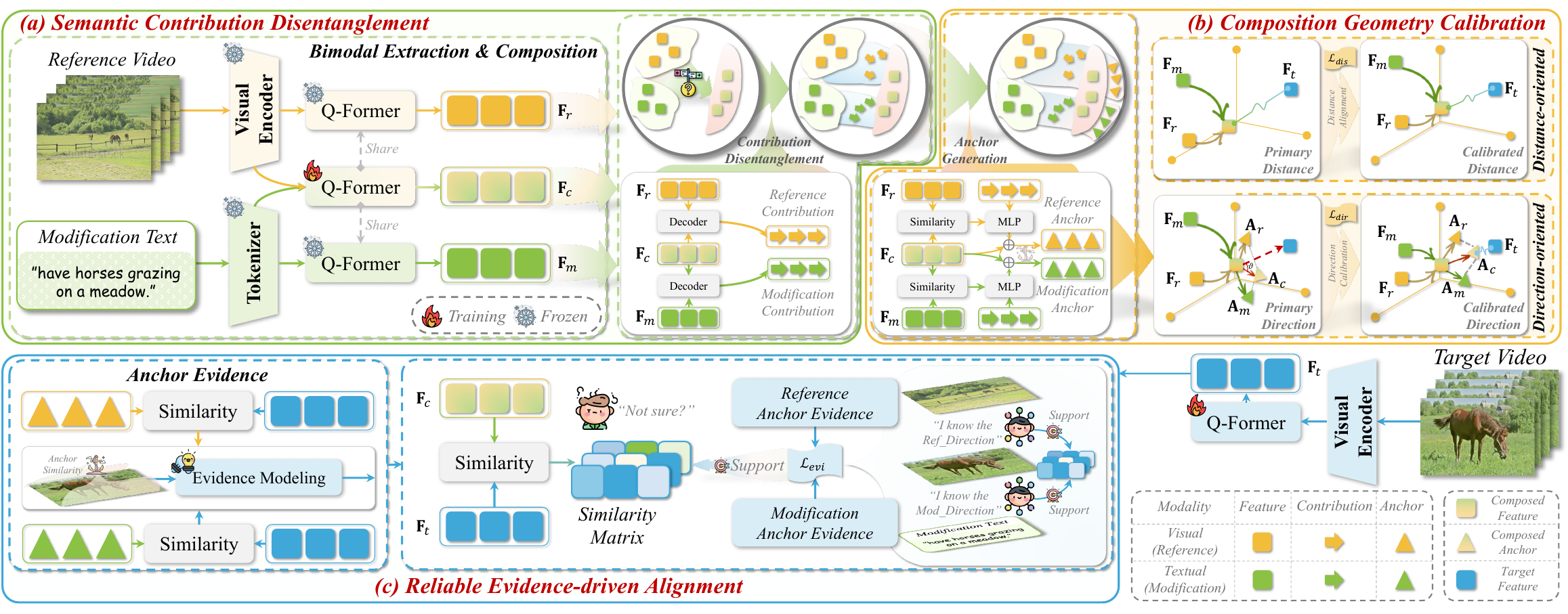}
\end{center}
   \caption{The proposed ReTrack consists of three key modules: (a) Semantic Contribution Disentanglement, (b) Composition Geometry Calibration, and (c) Reliable Evidence-driven Alignment.}
\label{fig:overview}
\end{figure*}

\section{Related Work}
Our work is closely related to Composed Video Retrieval (CVR) and Uncertainty Estimation.

\subsection{Composed Video Retrieval}
Similar to CIR~\cite{TME,HABIT,INTENT,HINT,MELT}, the CVR task focuses on developing models that interpret user-modified descriptions and reference videos for multimodal video retrieval. \textit{Ventura et al.}\cite{covr, covr-2} first formalized CVR and demonstrated the effectiveness of pretrained visual-linguistic models like BLIP\cite{blip} and BLIP-2~\cite{blip-2} for multimodal query understanding~\cite{li2025set,gao2024eraseanything,BML,he2024robust,ge2025gen4track}, adapting them to CVR with simple composition mechanisms. \textit{Thawakar et al.}\cite{covr_enrich} later enhanced query semantics with enriched captions. However, prior approaches overlook directional bias in composed features and the challenge of multiple similar candidates, leading to retrieval inaccuracies. ReTrack addresses these issues by calibrating feature bias and using directional anchors to compute bidirectional evidence, improving similarity estimation and retrieval accuracy.

\subsection{Uncertainty Estimation}
To quantify prediction uncertainty in deep neural networks~\cite{song1,Feng2024NoiseBox,zhou2024information,zhang2026towards,lan2025multi,qiu2025duet,Feng2023MaskCon,duan2025copinn,wang2024computing,yuan2025deep,Feng2022SSR}, much research~\cite{qiu2024tfb,zhang2023multi,yu2025cotextor,wang2025ascd,ge2024consistencies,ge2025beyond,wang2025r1trackdirectapplicationmllms} has focused on uncertainty estimation. Early methods used Bayesian theory, approximating posterior predictive distributions~\cite{baye-1}, leading to Bayesian Neural Networks (BNNs). However, BNNs suffer from high computational costs and slow inference for the deep learning technical~\cite{zhang2024fast,li2026multiple,gu2025mocount,li2025chatmotion,jia2024adaptive,liu2024graph,song4,ni2025wonderturbo,zhong2026collaborativemultiagentscriptsgeneration,ye2026gigaworld,wu2025spatiotemporal}. Evidential Deep Learning (EDL) addresses these limitations by modeling uncertainty through network outputs, achieving success in vision~\cite{yu2025yielding,zhou2023fastpillars,liao2025convex,pillarhist,song6,liu2023regformer,anonymous2025comptrack,zhang2026decoding,liu2025difflow3d,song13,liu2024dvlo,zhang2024cf,wang2026eeo,zhao2024balf,yu2025visual,zhou2024lidarptq,liu2024difflow3d,song5,liao2024globalpointer,focustrack,du2024self} and multi-modal tasks~\cite{EDL,sun2025roll,ge2025debate,jia2026ramrecover3dhuman,li2025human}. \textit{Sensoy et al.}~\cite{EDL} introduced subjective logic for improved uncertainty estimation and robustness, while \textit{Han et al.}~\cite{uncertainty-1} extended these ideas to multi-view classification with dynamic evidence fusion, enhancing reliability.
Inspired by EDL, ReTrack leverages bidirectional evidence interactions between directional anchors and target features. By adaptively weighting high-confidence samples, ReTrack more accurately aligns composite and target features, reducing the impact of similar candidate videos during retrieval.

\section{ReTrack}
As a key innovation, we introduce the ReTrack model, which calibrates directional bias in composed features and enables reliable composed-to-target similarity computation using evidence from calibrated directional anchors. As illustrated in Figure~\ref{fig:overview}, ReTrack comprises three key modules:
\textit{(a) Semantic Contribution Disentanglement,} which disentangles visual and textual contributions within composed features to support effective bias calibration (Section~\ref{SCD});
\textit{(b) Composition Geometry Calibration,} which constructs directional anchors from modal semantic contributions and reconstructs the composed features to calibrate directional bias (Section~\ref{CGC});
\textit{(c) Reliable Evidence-driven Alignment,} which uses bidirectional evidence between directional anchors and target features to weight high-credibility samples and reliably align composed features with targets (Section~\ref{REA}).
In this section, we first formalize the CVR task and then elaborate on each module of ReTrack.

\subsection{Problem Formulation}
The Composed Video Retrieval (CVR) task aims to retrieve the target video that fulfills the multimodal query. Let $\mathcal{T}\!\!=\!\!\left\{\left(x_{r},x_{m}, x_{t}\right)_{n}\right\}_{n=1}^{N}$ denote a set of $N$ triplets, where $x_{r}, x_{m}$ and $x_{t}$ refer to the reference video, modification text and target video, respectively. Our goal is to optimize a metric space where the embedding of the multimodal query ($x_{r}, x_{m}$) should be as close as possible to the corresponding target video $x_{t}$, formulated as, $\mathcal{G}\left(x_r, x_m\right) \rightarrow \mathcal{G}\left(x_t\right),$
where $\mathcal{G}$ denotes the to-be-optimized embedding function for both the multimodal query and target video.

\subsection{Semantic Contribution Disentanglement}
\label{SCD}
To calibrate directional bias in the composed feature, we first disentangle the semantic contributions of each modality. To this end, we introduce \textit{Semantic Contribution Disentanglement}, which first extracts features for the reference video, modification text, and their composed feature. It then separately interacts the composed feature with reference video and modification text branches, disentangling their respective semantic contributions. The disentanglement forms basis for subsequent directional calibration, as detailed below.

\noindent
\textbf{Bimodal Extraction \& Composition.}
Following previous work~\cite{covr-2,sprc,FineCIR}, we first leverage the Q-Former to extract the features of the reference video and the modification text, as well as their cross-modal composed feature, formulated as follows,
\begin{equation}
\left\{
\begin{aligned}
    &\textbf{F}_r=\operatorname{Q-Former}(\varPhi_\mathbb{I}(x_r)), \textbf{F}_m=\operatorname{Q-Former}(\varPhi_\mathbb{T}(x_m)),\\
    &\textbf{F}_c=\operatorname{Q-Former}(\varPhi_\mathbb{I}(x_r),\varPhi_\mathbb{T}(x_m)),
\end{aligned}
\label{Q-Former}
\right.
\end{equation}
where $\mathbf{F}_r, \mathbf{F}_m, \mathbf{F}_c\!\in\! \mathbb{R}^{Q\times D}$ are the reference feature, modification feature, and composed feature, respectively. $Q$ is the number of learnable queries for $N_f$ sampled frames, $N_f$ is the frame number, and $D$ is the embedding dimension. $\varPhi_{\mathbb{I}}$ and $\varPhi_{\mathbb{T}}$ denote the visual encoder and text tokenizer, respectively. Subsequently, for the target video, we apply the same manner to obtain the target feature $\mathbf{F}_t \in \mathbb{R}^{Q\times D}$.

\noindent
\textbf{Contribution Disentanglement.}
Subsequently, we disentangle the semantic contributions from the reference video and the modification text within the composed feature separately. Below, we take the reference video as an example.

To disentangle the semantic contribution of the reference video, we might naturally consider subtracting the modification text feature from the composed feature. However, this naive subtraction fails to capture the true visual semantic contribution due to the complexity of modification semantics. Thus, we introduce a \textit{Transformer Decoder} to more accurately extract the reference video's semantic contribution. Specifically, the reference video feature $\mathbf{F}_r$ is used as the \textit{Query}, and the composed feature $\mathbf{F}_c$ serves as both the \textit{Key} and \textit{Value}, yielding the reference semantic contribution $\mathbf{P}_r$,
\begin{equation}
\textbf{P}_r=\operatorname{Decoder}(Q=\textbf{F}_r, \{K,V\}=\textbf{F}_c),
    \label{TD}
\end{equation}
where $\mathbf{P}_r \in \mathbb{R}^{Q\times D}$ is the reference contribution. Similarly, we obtain the modification contribution $\mathbf{P}_m \in \mathbb{R}^{Q\times D}$.

\subsection{Composition Geometry Calibration}
\label{CGC}
To calibrate potential directional bias in the composed feature (i.e., an excessive bias toward the visual or textual modality at the expense of the other), we introduce the \textit{Composition Geometry Calibration} module, ensuring the calibrated feature remains close to the target video. This module first uses modal semantic contributions to generate bimodal directional anchors for each channel of the composed feature. It then employs \textit{Distance-oriented Alignment} to minimize the distance to the target feature, and \textit{Direction-oriented Calibration} to reconstruct the composed feature from these anchors, optimizing its direction relative to the target. This approach enables more accurate multimodal feature composition, as detailed below.

\noindent
\textbf{Anchor Generation.}
Firstly, since not all channels in the bimodal feature equally influence the composed direction, channels with greater compositional relations to the composed feature should be weighted more heavily in directional calibration. To address this, we introduce composition-oriented bimodal directional anchors based on semantic contributions. The computation of the reference anchor is described below as an example.

Specifically, we first introduce \textit{Point Weights} $\mathbf{W}_p \in \mathbb{R}^{Q\times D}$, which adaptively learn the weight of each channel feature's influence on directionality based on the similarity between the reference and composed feature, formulated as,
\begin{equation}
    \textbf{W}_p=\operatorname{MLP}(\textbf{F}_c\cdot{\textbf{F}_r}^\top).
    \label{point_w_}
\end{equation}

Subsequently, we leverage the point weights $\mathbf{W}_p$ to adjust the contribution of different feature channels in the semantic contributions to the composed direction, thereby generating the reference anchor $\mathbf{A}_r$, formulated as follows,
\begin{equation}
\textbf{A}_r=\textbf{F}_c+\textbf{W}_p\odot\textbf{P}_r,
    \label{point_w}
\end{equation}
where $\mathbf{A}_r \in \mathbb{R}^{Q\times D}$. Similarly, we obtain the modification anchor $\mathbf{A}_m \in \mathbb{R}^{Q\times D}$.

\noindent
\textbf{Distance-oriented Alignment.}
Secondly, to provide a more accurate distance basis for the subsequent calibration, we perform \textit{Distance-oriented Alignment}. In this part, we leverage a batch-based classification loss, which is widely utilized in CVR/CIR tasks~\cite{covr-2,sprc}, to pull the position of the composed feature closer to that of the target feature, formulated as follows,
\begin{equation}
    \mathcal{L}_{dis} \!=\! \frac{1}{B} \!\sum_{i=1}^{B} \!-\!\log\! \left\{ \frac{\exp \left\{  \mathcal{S}\left(\textbf{F}_{ci}, \textbf{F}_{ti}\right)  / \tau\right\}}{ \sum_{j=1}^{B} \exp \left\{ \mathcal{S}\left(\textbf{F}_{ci}, \textbf{F}_{tj}\right) / \tau \right\}  } \right\},
    \label{bbc}
\end{equation}
where $\mathcal{S}(\cdot, \cdot)$ is the similarity function, $B$ is the batch size, and $\tau$ is the temperature coefficient. $\textbf{F}_{ci}$ and $\textbf{F}_{ti}$ denote the \(i\)-th compose feature and target feature in the batch.

\begin{table*}[ht!]
  \centering
  \tabcolsep=14pt
\resizebox{0.75\linewidth}{!}{
    \begin{tabular}{l|cccc|c}
    \hline
    \hline
    \multicolumn{1}{c|}{\multirow{3}{*}{Method}} & \multicolumn{5}{c}{WebVid-CoVR-Test}\\
\cline{2-6}          & \multicolumn{4}{c|}{R@$k$}      & \multirow{2}{*}{Avg.}  \\
\cline{2-5}      & $k$=$1$   & $k$=$5$   & $k$=$10$  & $k$=$50$  &    \\
    \hline
    \multicolumn{6}{c}{\textbf{\textit{\underline{Pre-trianed Models}}}}\\
    CLIP~\cite{clip} & 44.37 & 69.13 & 77.62 & 93.00 & 71.03  \\
    BLIP~\cite{blip} & 45.46 & 70.46 & 79.54 & 93.27 & 72.18\\
    \hline
    \multicolumn{6}{c}{\textbf{\underline{\textit{CVR Models}}}}\\
    CoVR~\cite{covr} & 53.13 & 79.93 & 86.85 & 97.69 & 79.40  \\
    CoVR\_Enrich~\cite{covr_enrich} &\underline{60.12} & \underline{84.32} & 91.27 & \underline{98.72} & \underline{83.61}  \\
    CoVR-2~\cite{covr-2} & 59.82 & 83.84 & \underline{91.28} & 98.24 & 83.30 \\
    FDCA~\cite{fdca} & 54.80 & 82.27 & 89.84 & 97.70 & 81.15  \\
    \hline
    \textbf{ReTrack (Ours)} & \textbf{63.85} & \textbf{87.05} & \textbf{92.80} & \textbf{99.10} & \textbf{85.70}  \\
    \hline
    \hline
    \end{tabular}%
    }
  \caption{Performance comparison on the test set of the CVR dataset, WebVid-CoVR, relative to R@$k$($\%$). The overall best results are in bold, while the best results over baselines are underlined.}
  \label{tab:cvr}%
\end{table*}%

\noindent
\textbf{Direction-oriented Calibration}
Finally, we start from the directional anchors and impose their semantic contributions onto the composed feature to derive the composition directional anchor. We then use this composition directional anchor as an intermediary, pulling it closer to the target feature, thereby ensuring the accuracy of each modality's semantic contribution within the composed feature.

Specifically, we construct the composition directional anchor $\mathbf{A}_c \in \mathbb{R}^{Q\times D}$ based on the ``parallelogram law'' as,
\begin{equation}
    \textbf{A}_{c} \!=\! \left( \textbf{A}_r - \textbf{F}_c \right) + \left( \textbf{A}_m - \textbf{F}_c \right).
    \label{anchor_com}
\end{equation}

Subsequently, we compute the true directional vector from the original composed feature to the target feature as $\mathbf{A}_t = (\mathbf{F}_t - \mathbf{F}_c) \in \mathbb{R}^{Q\times D},$ which serves to guide the calibration of the composition directional anchor $\mathbf{A}_c$ toward the target feature, thereby eliminating directional bias and ensuring that the composition process more precisely points to the target feature in spatial direction, formulated as follows,
\begin{equation}
    \mathcal{L}_{dir} \!=\! \frac{1}{B} \!\sum_{i=1}^{B} \!-\!\log\! \left\{ \frac{\exp \left\{  \mathcal{S}\left(\textbf{A}_{ci}, \textbf{A}_{ti}\right)  / \tau\right\}}{ \sum_{j=1}^{B} \exp \left\{ \mathcal{S}\left(\textbf{A}_{ci}, \textbf{A}_{tj}\right) / \tau \right\}  } \right\},
    \label{dir}
\end{equation}
where $\mathcal{S}(\cdot, \cdot)$ is the similarity function, $B$ is the batch size, and $\tau$ is the temperature coefficient. $\mathbf{A}_{c_i}$ and $\mathbf{A}_{t_i}$ denote the $i$-th composition directional anchor and true directional vector in the batch, respectively.

\subsection{Reliable Evidence-driven Alignment}
\label{REA}
To reduce ReTrack's uncertainty when encountering similar candidate videos, we propose \textit{Reliable Evidence-driven Alignment}. This approach computes bidirectional evidence by interacting directional anchors with the target feature, automatically weights highly credible samples, and reliably aligns the composed feature with the target feature.

\noindent
\textbf{Evidence Modeling.}
To reduce the uncertainty in the alignment between the composed feature and the target feature, we utilize the \textit{Dempster-Shafer Theory of Evidence (DST)}~\cite{dst}. This theory is widely applied to handle available evidence from different sources in order to quantify the reliability of a given hypothesis. In our ReTrack model, we leverage DST to measure  correlation reliability between the two sets of directional anchors and the target feature, thereby further enhancing the reliability of the similarity matrix during the alignment process. In the following, we illustrate the evidence computation process using the reference anchor as an example.

Specifically, we first define the evidence vector in DST as $\mathbf{E} = [e_1, \dots, e_Q] \in \mathbb{R}^Q,$ which represents the matching evidence between each channel of the reference anchor $\mathbf{A}_r$ and the target feature $\mathbf{F}_t$. Following \textit{Evidence Deep
Learning (EDL)}~\cite{EDL}, we utilize the \textit{Subjective Logic} to formulate the evidence as follows,
\begin{equation}
    e_q=\operatorname{exp}(\max^Q_{\hat{q}=1}\left(\textbf{A}_{r(q)}\cdot \textbf{F}_{t}^\top\right)_{\hat{q}} /{\tau}),
    \label{evidence_c}
\end{equation}
where $Q$ is the number of learnable queries in the Q-Former, and $\mathbf{A}_{r(q)}$ denotes the $q$-th channel of the reference anchor. $e_q$ is the matching evidence between the $q$-th channel of the reference anchor and the target feature. Based on the matching evidence from all channels, we further compute the belief mass for each channel to measure each channel's confidence in its own decision, formulated as follows,
\begin{equation}
    b_q = \frac{e_q}{\sum^Q_{\hat{q}=1}\left(e_{\hat{q}}+1\right)}.
    \label{belief-b}
\end{equation}

Based on each channel's belief mass of its own decision, we can derive the overall correlation reliability of the reference anchor, which denotes the directional semantic information during the composition process, formulated as,
\begin{equation}
    \mathbb{E}_r = \sum^{Q}_{q=1}b_q=1-\frac{Q}{\sum^Q_{\hat{q}=1}\left(e_{\hat{q}}+1\right)}.
    \label{belief_e}
\end{equation}

In the same manner, we can obtain the correlation reliability between the directional semantic information of the modification text and the target feature, denoted as $\mathbb{E}_m$.

\noindent
\textbf{Optimization.}
Subsequently, based on EDL~\cite{EDL}, we argue that the correlation reliability $\mathbb{E}_r,\mathbb{E}_m$ should be positively correlated with the similarity between the composed feature and the target feature within the batch, for better comprehension.
Thus, based on the two sets of correlation reliability, we design an evidence-driven regularization loss to ensure consistency between the similarity measurement and the correlation reliability, thereby enhancing the reliability of the similarity between the composed feature and the target feature, formulated as,

\begin{equation}
    \mathcal{L}_{evi}=\frac{1}{B}\sum^B_{b=1}\left(\mathbb{E}_{rb}\!\!-\!\! \mathcal{S}\left(\textbf{F}_{cb}, \textbf{F}_{tb}\right) \right)^2\!\!+\!\!
    \left(\mathbb{E}_{mb}\!\!-\!\!\mathcal{S}\left(\textbf{F}_{cb}, \textbf{F}_{tb}\right) \right)^2,
    \label{belief}
\end{equation}
where $B$ is the batch size, and $\mathbf{F}_{cb}, \mathbf{F}_{tb}$ denote the $b$-th composed feature and target feature in the batch, respectively.

Finally, we obtain the final loss function for ReTrack as,
\begin{equation}
\mathbf{\Theta^{*}}=
\underset{\mathbf{\Theta}}{\arg \min } \left( \mathcal{L}_{dis} + \kappa \mathcal{L}_{dir} + \lambda \mathcal{L}_{evi}\right),
\label{optimization}
\end{equation}
where $\mathbf{\Theta}$ is the ReTrack parameter to be learned and $\kappa,\lambda$ are the trade-off hyper-parameters.

\begin{table*}[ht]
  \centering
  \tabcolsep=2.8pt
\resizebox{0.9\linewidth}{!}{
    \begin{tabular}{l|cc|cc|cc|cccc|ccc}
    \hline
    \hline
    \multicolumn{1}{c|}{\multirow{3}{*}{Method}} & \multicolumn{6}{c|}{FashionIQ} & \multicolumn{7}{c}{CIRR} \\
    \cline{2-14}
        
    & \multicolumn{2}{c|}{Dresses} & \multicolumn{2}{c|}{Shirts} & \multicolumn{2}{c|}{Tops\&Tees} & \multicolumn{4}{c|}{R@$k$} & \multicolumn{3}{c}{R$_{\text{sub}}$@$k$} \\
    \cline{2-14}
    & R@10 & R@50 & R@10 & R@50 & R@10 & R@50 & $k$=1 & $k$=5 & $k$=10 & $k$=50 & $k$=1 & $k$=2 & $k$=3 \\
    \hline
    \multicolumn{14}{c}{\textbf{\textit{\underline{CIR Models}}}}\\
    TG-CIR~\cite{tgcir} & 45.22  & 69.66  & 52.60  & 72.52  & 56.14  & 77.10  & 45.25  & 78.29  & 87.16  & 97.30  & 72.84  & 89.25  & 95.13  \\
    SSN~\cite{ssn} & 34.36  & 60.78  & 38.13  & 61.83  & 44.26  & 69.05  & 43.91  & 77.25  & 86.48  & 97.45  & 71.76  & 88.63  & 95.54  \\
    SADN~\cite{sadn} & 40.01  & 65.10  & 43.67  & 66.05  & 48.04  & 70.93  & 44.27  & 78.10  & 87.71 & 97.89 & 72.34 & 88.70  & 95.23 \\
    SPRC~\cite{sprc} & 49.18  & 72.43  & 55.64  & 73.89  & 59.35  & 78.58  & \underline{51.96} & \underline{82.12} & \underline{89.74} & 97.69 & \textbf{80.65} & \underline{92.31} & \underline{96.60} \\
    LIMN~\cite{limn} & 50.72  & 74.52  & 56.08  & 77.09  & 60.94  & 81.85  & 43.64  & 75.37  & 85.42  & 97.04  & 69.01  & 86.22  & 94.19  \\
    LIMN+~\cite{limn} & \underline{52.11}  & \underline{75.21}  & \underline{57.51}  & \underline{77.92}  & \underline{62.67}  & \underline{82.66}  & 43.33  & 75.41  & 85.81  & 97.21  & 69.28  & 86.43  & 94.26  \\
    IUDC~\cite{iudc} & 35.22  & 61.90  & 41.86  & 63.52  & 42.19  & 69.23  & -     & -     & -     & -     & -     & -     & - \\
    ENCODER~\cite{encoder} & 51.51  & 76.95  & 54.86  & 74.93  & 62.01  & 80.88  & 46.10  & 77.98  & 87.16  & 97.64  & 76.92  & 90.41  & 95.95  \\
    \multicolumn{14}{c}{\textbf{\textit{\underline{CVR Models}}}}\\
    CoVR~\cite{covr} & 44.55  & 69.03  & 48.43  & 67.42  & 52.60  & 74.31  & 49.69  & 78.60  & 86.77  & 94.31  & 75.01  & 88.12  & 93.16  \\
    CoVR\_Enrich~\cite{covr_enrich} & 46.12  & 69.52  & 49.61  & 68.88  & 53.79  & 74.74  & 51.03  & -     & 88.93  & 97.53  & 76.51  & -     & 95.76  \\
    CoVR-2~\cite{covr-2} & 46.53  & 69.60  & 51.23  & 70.64  & 52.14  & 73.27  & 50.43 & 81.08 & 88.89 & \underline{98.05} & 76.75 & 90.34 & 95.78 \\
    \hline
    \textbf{ReTrack (Ours)} & \textbf{52.91} & \textbf{77.54} & \textbf{61.91} & \textbf{81.26} & \textbf{63.22} & \textbf{83.36} & \textbf{52.34} & \textbf{82.53} & \textbf{90.34} & \textbf{98.13} & \underline{79.64} & \textbf{92.58} & \textbf{96.99} \\
    \hline
    \hline
    \end{tabular}
    }
      \caption{Performance comparison on the CIR dataset, FashionIQ and CIRR, relative to R@$k$(\%). The overall best results are in bold, while the best results over baselines are underlined.}
  \label{tab:cir}
\end{table*}

\section{Experiments}
This section delves into our comprehensive experiments of ReTrack and the corresponding analyses.

\subsection{Experimental Setup}

\textbf{Datasets.} To comprehensively evaluate the efficacy and generalizability of the proposed ReTrack, we conduct experiments on both CVR and CIR tasks. For the CVR task, we adopt the large-scale open-domain WebVid-CoVR~\cite{covr}. For the CIR task, we employ the widely used fashion-domain FashionIQ dataset~\cite{FashionIQ}, and the open-domain CIRR dataset~\cite{cirr}.

\noindent
\textbf{Evaluation Metrics.} To ensure fair comparisons, we follow the standard evaluation protocols of each dataset and report Recall@$k$ (R@$k$) as the primary metric: \textbf{1) WebVid-CoVR:} R@$\{1,\!5,\!10,\!50\}$, along with their mean. \textbf{2) FashionIQ:} R@$\{10,50\}$ for each category. \textbf{3) CIRR:} R@$\{1,5,10,50\}$, and subset-based metrics R$_{\text{sub}}$@\{$1,2,3$\}.

\noindent
\textbf{Implementation Details.} Following previous works~\cite{covr-2}, we adopt BLIP-2~\cite{blip-2} fine-tuned on the COCO dataset with $364$-pixel input resolution as the backbone model for ReTrack and freeze the ViT during training. The frame number $N_f=4$ and the number of learning query $Q=32N_f$.
For the trade-off hyper-parameters in Eq.(\ref{optimization}), we conduct a grid search to set $\lambda=1.0$ and $\kappa=0.5$. The temperature coefficient $\tau=0.1$.
ReTrack is trained with a batch size of $64$ using the AdamW optimizer with a learning rate of $2e-5$. Training is performed for $5,10$ epochs on CVR and CIR datasets. 
All experiments are conducted on an NVIDIA V100 GPU with $32$GB memory.

\subsection{Performance Comparison}
To validate the performance and generalization of ReTrack, we conduct extensive comparisons on CVR and CIR tasks.

\noindent
\textit{\textbf{On CVR Task.}}
As shown in Table~\ref{tab:cvr}, we compare two categories of baselines: pretrained models and CVR models. The results reveal the following observations:
\textbf{1)} ReTrack achieves the best performance across all evaluation metrics on both CVR datasets. Specifically, on WebVid-CoVR, ReTrack yields a $2.50$\% improvement in the mean metric. And the R$1$ metric improves significantly. This demonstrates that by calibrating directional bias in the composed feature and enhancing the reliability of the similarity between the composed feature and the target feature, ReTrack effectively improves its understanding of multi-modal queries.
\textbf{2)} CoVR\_Enrich performs sub-optimal on WebVid-CoVR, likely due to its use of extra generated captions to improve cross-modal perception. In contrast, ReTrack surpasses it without extra inputs, relying solely on Composition Geometry Calibration and Reliable Evidence-driven Alignment.

\noindent
\textit{\textbf{On CIR Task.}}
As shown in Table~\ref{tab:cir}, we compare CIR models and CVR models. The results yield the following key insights:
\textbf{1)} ReTrack achieves the best performance on all metrics across both CIR datasets. Compared to the second-best method, ReTrack attains relative improvements of $1.54$\%, $7.7$\%, and $0.88$\% in R@$10$ on FashionIQ for different categories, and $0.73$\% in R@$1$ on CIRR. This demonstrates that ReTrack's multimodal semantic disentanglement and calibration-based feature modeling provide strong domain generalization.
\textbf{2)} Most CVR models lag behind specialized CIR models on CIR tasks, likely due to their focus on global visual perspectives and reliance on repeated key targets across frames, which can overlook single-frame visual details and introduce semantic bias. In contrast, ReTrack effectively attends to multimodal details and performs cross-modal calibration, enabling precise semantic composition for both CVR and CIR. This highlights ReTrack's strong generalization in visual-modality semantic understanding.

\subsection{Ablation Study}
To assess the effect of each ReTrack module, we perform detailed ablation studies across the following variant groups:
\textit{\textbf{G[A]: Ablation on Semantic Contribution Disentanglement}}
\textbf{D\#(1) wo\_C\_ref}, \textbf{D\#(2) wo\_C\_mod}: Remove the semantic contribution from the reference video or modification text, respectively, using only one modality's contribution.
\textbf{D\#(3) wo\_SCD}: Remove \textit{Semantic Contribution Disentanglement} and use original features instead.
\textit{\textbf{G[B]: Ablation on Composition Geometry Calibration}}
\textbf{D\#(4) wo\_$\mathcal{L}{dis}$}: Remove \textit{Distance-oriented Alignment} to test its positional role in calibration.
\textbf{D\#(5) wo\_A\_ref}, \textbf{D\#(6) wo\_A\_mod}: Remove the reference or modification anchor in Eq.(\ref{anchor_com}), respectively.
\textbf{D\#(7) wo\_$\mathcal{L}_{dir}$}: Remove \textit{Direction-oriented Calibration} $\mathcal{L}_{dir}$ in Eq.(\ref{dir}).
\textit{\textbf{G[C]: Ablation on Reliable Evidence-driven Alignment}}
\textbf{D\#(8) wo\_Evi\_ref}, \textbf{D\#(9) wo\_Evi\_mod}: Remove the reference or modification evidence terms from the regularization loss.
\textbf{D\#(10) wo\_$\mathcal{L}_{evi}$}: Remove the entire evidence-driven regularization loss.
\textit{\textbf{G[D]: Ablation on Evidence Calculation}}
\textbf{D\#(11) w\_RELU}, \textbf{D\#(12) w\_Softplus}: Replace the exponential evidence computation in Eq.~(\ref{evidence_c}) with ReLU and Softplus, to test the function choice.

\begin{table}[ht]
  \centering
  \small
    \begin{tabular}{c|c|cc|c|c}
    \hline
    \hline
    \multicolumn{1}{c|}{\multirow{2}{*}{\textbf{D\#}}} & \multirow{2}{*}{\textbf{Derivatives}} & \multicolumn{2}{c|}{\textbf{FIQ-Avg.}} & \multicolumn{1}{c|}{\textbf{CIRR}} & \multicolumn{1}{c}{\textbf{WebVid}} \\
\cline{3-6}    \multicolumn{1}{c|}{} &       & \textbf{R@10} & \textbf{R@50} & \textbf{Avg.} & \textbf{Avg.}\\
    \hline
    \multicolumn{6}{c}{\textit{\textbf{G[A]: Semantic Contribution Disentanglement}}} \\
    \hline
    1     & wo\_C\_ref & 58.84 & 80.25 & 79.86 & 83.90  \\
    2 & wo\_C\_mod & 58.68 & 79.94 & 79.54 & 84.20\\
    3 & wo\_SCD & 57.69 & 78.48 & 78.49 & 83.37 \\
    \hline
    \multicolumn{6}{c}{\textit{\textbf{G[B]: Composition Geometry Calibration}}} \\
    \hline
    4     & wo\_$\mathcal{L}_{dis}$ & 3.78 & 9.12 & 16.08 & 27.59\\
    5     & wo\_A\_ref & 58.21 & 79.48 & 79.73 & 84.33  \\
    6 & wo\_A\_mod & 58.31 & 79.87 & 79.54 & 84.27  \\
    7 & wo\_$\mathcal{L}_{dir}$ & 57.64 & 78.82 & 79.68 & 83.66 \\

    \hline
    \multicolumn{6}{c}{\textit{\textbf{G[C]: Reliable Evidence-driven Alignment}}} \\
    \hline
    8& wo\_Evi\_ref & 59.11 & 80.09 & 80.03 & 84.27 \\
    9 & wo\_Evi\_mod & 59.03 & 80.08 & 80.59 & 84.19 \\
    10    & wo\_$\mathcal{L}_{evi}$ & 56.93 & 78.19 & 78.94 & 83.02  \\
    \hline
    \multicolumn{6}{c}{\textit{\textbf{G[D]: Calculation of Evidence}}} \\
    \hline
    11 & w\_ReLU & 59.02 & 80.07 & 80.68 & 84.65  \\
    12 & w\_Softplus & 59.11 & 80.19 & 81.01 & 84.54  \\
    \hline
    \multicolumn{2}{c|}{\textbf{ReTrack (Ours)}} & \textbf{59.35} & \textbf{80.72} & \textbf{81.09} & \textbf{85.70} \\
    \hline
    \hline
    \end{tabular}%
      \caption{Ablation study on three CVR and CIR datasets.}
  \label{tab:Ablation}%
\end{table}%

From Table~\ref{tab:Ablation}, we obtain the following observations.
\textbf{1)} Compared to the full ReTrack model, \textbf{D\#(1)} and \textbf{D\#(2)} show slight performance drops, indicating the necessity of disentangling both visual and textual contributions for effective calibration and retrieval. 
\textbf{2)} Within G[A], \textbf{D\#(3)} shows the largest decline, indicating that jointing multimodal semantic bias is essential for calibrating modality-specific semantic deviations, which in turn enhances multimodal understanding.
\textbf{3)} \textbf{D\#(4)} yields a notable decrease, underscoring the importance of distance guidance for direction calibration. Both \textbf{D\#(5)} and \textbf{D\#(6)} reduce performance, confirming that reference and modification anchors each provide essential directional cues. \textbf{D\#(7)} exhibits an even greater drop, reinforcing the role of distance in measuring each modality's contribution.
\textbf{4)} \textbf{D\#(8)} and \textbf{D\#(9)} also lead to declines, showing that uncertainty quantification from both modalities is vital for reliable alignment. \textbf{D\#(10)} results in the sharpest drop in G[C], showing the importance of evidence-driven regularization for robust retrieval.
\textbf{5)} \textbf{D\#(11)} and \textbf{D\#(12)} examine evidence computation methods, which reveals that evidence-theory–compliant approaches can estimate data uncertainty, with the exponential method adopted as optimal.

\begin{figure}[h]
\begin{center}
\includegraphics[width=0.97\linewidth]{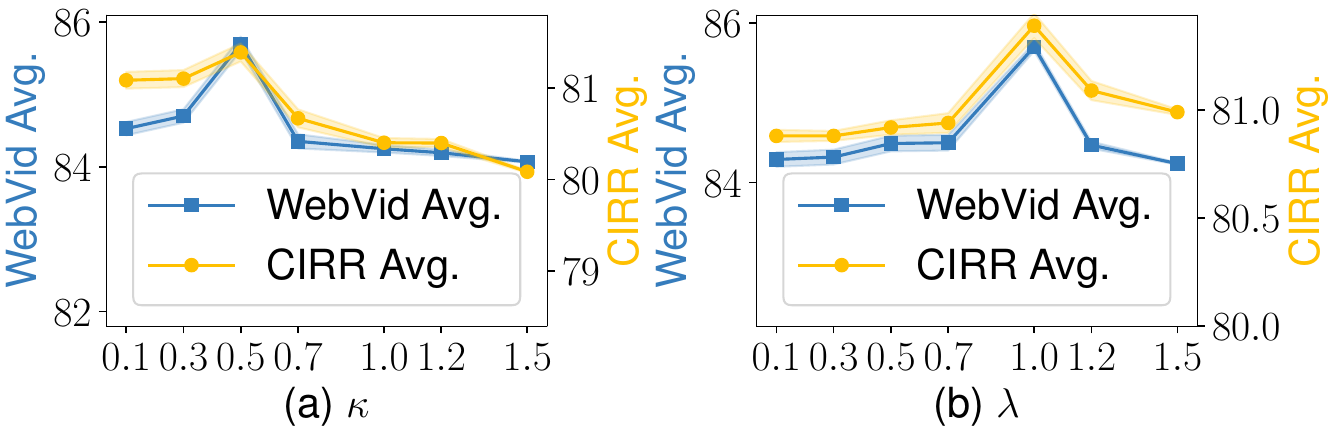}
\end{center}
   \caption{Sensitivity to the hyper-parameters (a) $\kappa$, and (b) $\lambda$ on WebVid-CoVR and CIRR datasets.}
\label{fig:sen}
\end{figure}

To analyse ReTrack's sensitivity to the hyperparameter $\kappa$ and $\lambda$ in Eq.(\ref{optimization}), we present results on WebVid-CoVR and CIRR in Figure~\ref{fig:sen}. We observe that, for both datasets, performance first increases and then decreases as $\kappa$ and $\lambda$ increase. This behavior is reasonable because the composition geometry requiring calibration does not exhibit unbounded deviation but lies within a limited range, so balanced hyperparameters are needed to constrain the degree of calibration. Moreover, a larger $\lambda$ effectively applies reliable evidence to the corresponding channels; however, not all channels require high evidence support, since some channels may inherently lack reliable semantic information. Thus, excessively large values lead to performance degradation.

\subsection{Case Study}
\begin{figure}[h]
\begin{center}
\includegraphics[width=\linewidth]{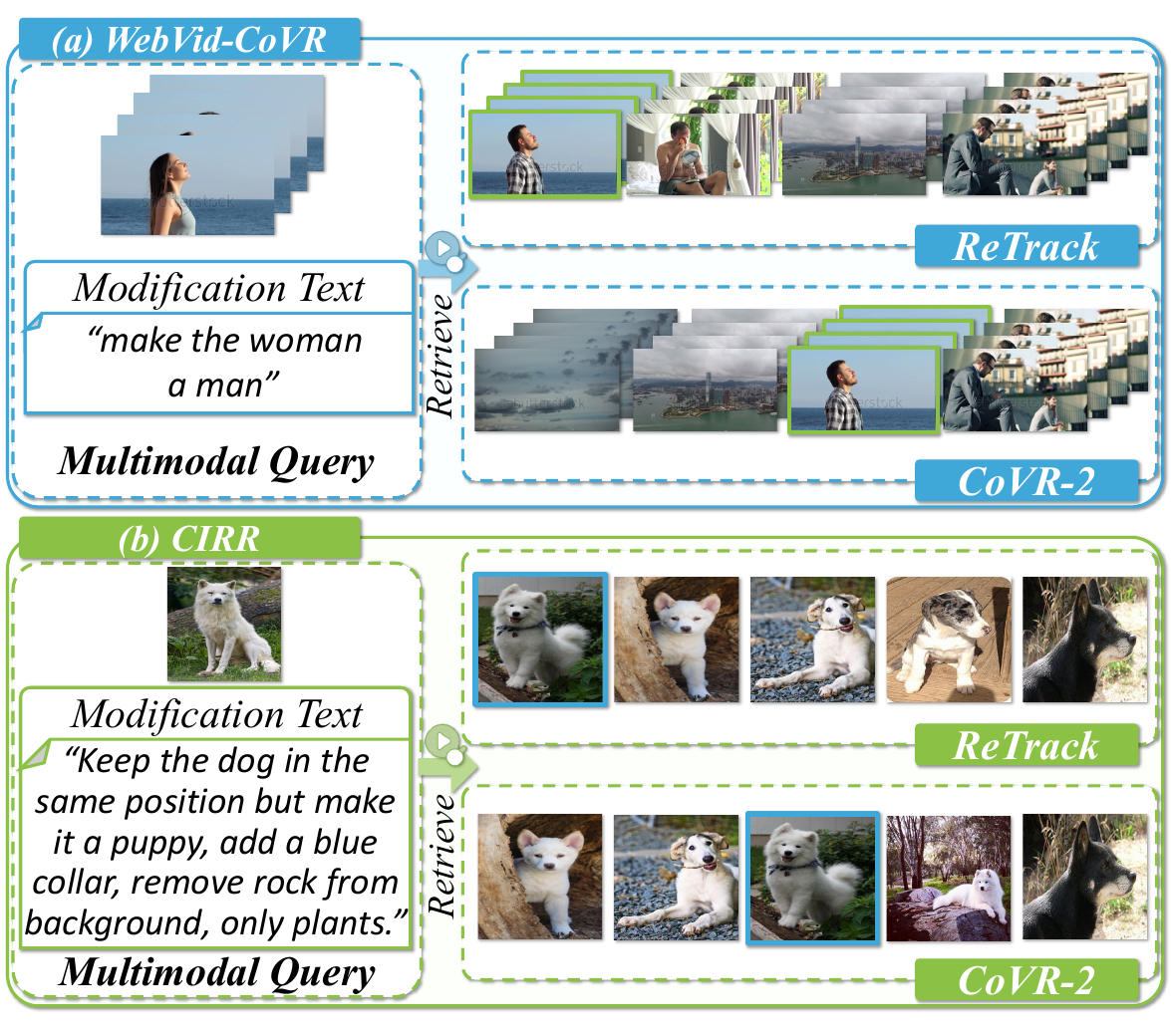}
\end{center}
   \caption{Case study on (a) WebVid-CoVR and (b) CIRR.}
\label{fig:case}
\end{figure}
As shown in Figure~\ref{fig:case}, we compare retrieval results from our ReTrack model and the representative CVR model CoVR-2 on WebVid-CoVR and CIRR, with the following observations:
\textbf{1)} In Figure~\ref{fig:case}(a), ReTrack retrieves the target video at rank~$1$, while CoVR-2 returns two ``sea-and-sky'' videos as its top results. The prevalence of ``sky'' and ``ocean'' in the reference video introduces high uncertainty in video semantics, reducing the text's contribution and resulting in CoVR-2's inaccurate retrieval. Additionally, CoVR-2's composed feature becomes overly text-biased due to the emphasis on ``man'' in the modification text. By leveraging evidence-driven uncertainty quantification, ReTrack effectively mitigates background semantic interference and achieves higher-quality results, demonstrating the value of its bias calibration and reliable similarity computation.
\textbf{2)} In Figure~\ref{fig:case}(b), ReTrack ranks the target image first, whereas CoVR-2 places it third. The modification text includes several requirements with low uncertainty, and ReTrack accurately captures these, yielding more complete matches. CoVR-2, in contrast, retrieves an image meeting only some requirements as rank $1$. This underscores the need for balanced modality contributions in forming the composed feature and reliable similarity computation.

\section{Conclusion}
In this work, we investigate the novel CVR task. Although previous methods have achieved impressive progress, they neglect the potential directional bias in the composed feature, which may lead to suboptimal retrieval performance. To address this limitation, we propose ReTrack, the first CVR framework that improves multi-modal query understanding by correcting directional bias in the composed feature. ReTrack calibrates the directional bias by computing modality-specific semantic contributions, and leverages the calibrated directional anchors to generate bidirectional evidence, enabling reliable composed-to-target similarity estimation. In addition, ReTrack is also compatible with CIR and achieves state-of-the-art performance on three benchmark datasets covering both CVR and CIR tasks. In future work, we plan to extend our method to multi-turn interactive Composed Multi-modal Retrieval.

\section*{Acknowledgments}
This work was supported in part by the National Natural Science Foundation of China, No.:62276155, No.:62576195, No.:62376140, and No.:U23A20315; and the Special Fund for Taishan Scholar Project of Shandong Province; in part by the China National University Student Innovation \& Entrepreneurship Development Program, No.:2025282 and No.:2025283.

 \clearpage

\appendix
\noindent
This is the supplementary material of the submitted paper \textit{\textbf{``ReTrack: Evidence‑Driven Dual‑Stream Directional Anchor Calibration Network for Composed Video Retrieval''}}. The content catalog is as follows:
\begin{itemize}
    \item \textbf{Appendix~\ref{appendix:datasets}}: Datasets
    \begin{itemize}
        \item \textbf{Appendix~\ref{appendix:CVR Datasets}}: CVR Datasets
        \item \textbf{Appendix~\ref{appendix:CIR Datasets}}: CIR Datasets
    \end{itemize}

    \item \textbf{Appendix~\ref{appendix:Derivation of Evidence Theory}}: Derivation of Evidence Theory
    \begin{itemize}
        \item \textbf{Appendix~\ref{Preliminary}}: Preliminary
        \item \textbf{Appendix~\ref{Construction of the Matching Hypothesis Space Based on DST}}: Construction of the Matching Hypothesis Space Based on DST
        \item \textbf{Appendix~\ref{Evidence Construction Based on Subjective Logic Theory}}: Evidence Construction Based on Subjective Logic Theory
    \end{itemize}

    \item \textbf{Appendix~\ref{appendix:Additional Performance Comparison}}: Additional Performance Comparison
    \begin{itemize}
        \item \textbf{Appendix~\ref{appendix:additional_performance_cir_cvr}}: Complete Performance Comparison on CIR and CVR Task
        \item \textbf{Appendix~\ref{appendix:additional_performance_efficiency}}: Efficiency Evaluation
    \end{itemize}

    \item \textbf{Appendix~\ref{appendix:algorithm}}: Algorithm of Retrack's Training Procedure

    \item \textbf{Appendix~\ref{appendix:case}}: More Case Study
\end{itemize}

\section{Datasets}
\label{appendix:datasets}

To comprehensively evaluate our proposed model, ReTrack, we selected three benchmark datasets, including one CVR dataset, WebVid-CoVR and two CIR datasets (FashionIQ and CIRR). The details of each dataset are provided below.

\subsection{CVR Datasets}
\label{appendix:CVR Datasets}

\begin{itemize}
    \item \textbf{WebVid-CoVR}  
    WebVid-CoVR is the first large-scale benchmark specifically designed for the CVR task. The dataset is derived from the WebVid-2M~\cite{Webvid-2M} dataset and contains approximately 1.6 million CVR triplets, spanning around 131k unique videos and 467k distinct modification texts. On average, each video has a duration of 16.8 seconds, and each modification text consists of approximately 4.8 words. Each target video is associated with roughly 12.7 triplets. The test set includes 2,500 high-quality triplets, carefully selected from the WebVid-10M dataset after an intensive annotation and noise removal process, providing a robust and challenging evaluation benchmark.
\end{itemize}

\subsection{CIR Datasets}
\label{appendix:CIR Datasets}

\begin{itemize}
    \item \textbf{FashionIQ}  
    FashionIQ is a dataset specifically designed for fashion-oriented image retrieval tasks. It consists of 77,684 online images, which are paired into 30,134 annotated triplets across three representative fashion categories: dresses, shirts, and T-shirts. The dataset is intended to evaluate multi-modal image retrieval capabilities in the fashion domain, focusing on the semantic relationship between images and modification texts.

    \item \textbf{CIRR}  
    CIRR is constructed from real-world scene images originating from the NLVR2 natural language visual reasoning dataset~\cite{NLVR2}. CIRR contains 36,554 annotated triplets and 21,552 images. Unlike FashionIQ, CIRR emphasizes the complex interactions among multiple objects in natural scenes, which helps mitigate the limitations of overfitting to a specific domain. Additionally, CIRR addresses the problem of incomplete annotations, which often leads to numerous false negatives in datasets like FashionIQ, and includes a specialized subset for fine-grained contrastive evaluation. CIRR is particularly suitable for evaluating models' performance in complex scenes involving object interactions and the integration of multi-modal data.
\end{itemize}

Through these datasets, we are able to evaluate ReTrack on both large-scale, web-sourced video data (WebVid-CoVR), demonstrating its versatility and robustness in handling diverse video retrieval scenarios. we are also able to evaluate ReTrack in both fashion-specific domains (FashionIQ) and more complex natural scene domains (CIRR). 

These datasets collectively provide a comprehensive evaluation framework for ReTrack, encompassing both CVR and CIR tasks across diverse domains.

\section{Derivation of Evidence Theory}
\label{appendix:Derivation of Evidence Theory}
\subsection{Preliminary}
\label{Preliminary}

\textbf{Dirichlet-based hypothesis probability estimation.} The Dirichlet distribution is commonly adopted as the conjugate prior of the multinomial distribution and is used to model the uncertainty associated with multiple evidence sources in the composed evidence process. Its probability density function is defined as follows,

\begin{equation}
p(\mathbf{x} \mid \boldsymbol{\alpha}) = \frac{1}{B(\boldsymbol{\alpha})} \prod_{i=1}^{K} x_i^{\alpha_i - 1},
\label{dirichlet_density}
\end{equation}
where $K$ denotes the number of categories, $x=(x_1,x_2,\ldots,x_K) $ is a probability vector representing the probability of each category, and $\alpha=(\alpha_1,\alpha_2,\ldots,\alpha_K) $ represents the confidence or prior knowledge associated with each category, where $\alpha_i$ corresponds to the confidence for the $i$-th category. $B(\alpha)$ is the Beta function used for normalization. Assuming that there are $Q$ evidence sources, the belief mass assigned by evidence source $m_q$ to hypothesis $A$ is denoted as $m_q(A)$, and these belief masses can be mapped to the Dirichlet distribution parameters $\alpha_i$ through the following process,

\begin{equation}
\alpha_i = \alpha_0 + \sum_{q=1}^{Q} m_q(A_i),
\label{alpha_update}
\end{equation}
where $\alpha_0$ is a constant, typically set to $1$, indicating an initial balanced belief. If multiple evidence sources support a particular hypothesis, their corresponding belief masses will be accumulated into $\alpha_i$. Once the Dirichlet distribution parameters $(\alpha_1,\alpha_2,\ldots,\alpha_K)$ are obtained, the distribution can be used to represent the belief mass probability distribution over each hypothesis. This distribution will be utilized in the subsequent theoretical derivations.

\subsection{Construction of the Matching Hypothesis Space Based on DST}
\label{Construction of the Matching Hypothesis Space Based on DST}
The Dempster-Shafer Theory of Evidence (DST) was proposed by Arthur Dempster and Glenn Shafer~\cite{dst}. It provides a theoretical framework for managing uncertainty, ambiguity, and conflicting evidence. The core idea of DST is to represent and combine evidence from multiple sources using a set-based mathematical structure, in order to perform rational reasoning and quantify the reliability of a given hypothesis. In our proposed \textbf{ReTrack} model, to reduce the uncertainty in the alignment process between the composed feature and the target feature, we introduce DST to compute reliable evidence between the two sets of directional anchors and the target feature. This enhances the reliability of the similarity matrix during alignment. We begin by introducing the hypothesis space of matching between composed features and target features, which underpins the formulation of DST. Subsequently, we decompose this structure to compute the reliable evidence between the directional anchors and the target feature.

\textbf{Matching hypothesis space between composed features and target features.} According to DST, we first define a hypothesis space $\Theta$, which contains all possible hypotheses. In our model, this corresponds to all possible matching combinations between the composed feature $F_c$ and the candidate target features $F_t$, where $Q$ is the number of learnable queries for $N_f$ sampled frames, $N_f$ is the frame number, and $D$ is the embedding dimension. The matching hypothesis space is formulated as follows,

\begin{equation}
\Theta = \{A_1, A_2, \ldots, A_N\},
\label{frame_of_discernment}
\end{equation}
where each hypothesis $A_i \in \Theta$ represents a possible matching configuration between the composed feature $F_c$ and the target feature $F_t$. By employing the Basic Probability Assignment (BPA) in DST, we can quantify the belief mass $m(A)$ associated with each hypothesis $A_i$, which reflects the degree to which the matching configuration in hypothesis $A$ is supported by the available evidence with respect to $F_c$. The BPA satisfies the following properties,

\begin{equation}
\sum_{A_i \subseteq \Theta} m(A_i) = 1 \quad \text{and} \quad m(A_i) \geq 0 \quad \text{for all } A_i \subseteq \Theta.
\label{bpa_constraint}
\end{equation}

The above formulation indicates that within a batch, each composed feature is assumed to have at least one corresponding target feature, thus $m(A_i) \geq 0$ and the sum of all hypothesis probabilities equals $1$.

\textbf{Matching hypothesis space between the two sets of directional anchors and the target feature.} According to DST, when multiple independent evidence sources are available, Dempster's rule is employed to fuse the evidence. In our approach, since the composed feature integrates semantic contributions from both visual and textual modalities, the original matching hypothesis space between the composed feature and the target feature is further decomposed into two separate matching spaces: one between the visual anchors and the target feature, and the other between the textual anchors and the target feature. Specifically, taking the directional anchors from the reference video as an example, the proposition space $\Theta$ is redefined as the set of all possible matching hypotheses between each anchor and the target feature. The degree of matching between the $Q$ channels and $F_t$ can be regarded as $Q$ independent evidence sources, as each channel encodes distinct semantic information. According to Dempster's rule, we perform evidence fusion over these $Q$ evidence sources, which is formulated as follows,

\begin{equation}
m_{\text{combined}}(A) = \frac{1}{1 - K} \sum_{B \cap C = A} \prod_{q=1}^{Q} m_q(B) \, m_q(C),
\label{dempster_combination}
\end{equation}
where $A$ is a hypothesis in the proposition space (for simplicity, we focus on a single composed feature and thus denote $A_i$ simply as $A$, representing the event that the reference video anchor matches the target feature). $B$ and $C$ are subsets of the proposition space supported by the evidence sources $m_q$. $K$ denotes the degree of conflict, which measures the level of disagreement among evidence sources. The computation of $K$ is given as follows,

\begin{equation}
K = \sum_{B \cap C = \varnothing} \prod_{q=1}^{Q} m_q(B) \, m_q(C).
\label{conflict_coefficient}
\end{equation}

The fused evidence $m_{\text{combined}}(A)$ represents the overall degree of support from all evidence sources (i.e., the matching measures of all channels) for the matching event between the reference video anchor and the target feature. Correspondingly, if $K = 1$, it indicates a complete conflict among the evidence sources, implying that no source supports the match between the reference video anchor and the target feature as described by hypothesis $A$, and thus the evidence cannot be fused. Therefore, by evaluating $m_q$, we can assess the reliability of hypothesis $A$, that is, the matching configuration between the composed feature $F_c$ and the target feature $F_t$. In the following section, we elaborate on the evaluation process based on $m_q$ and the construction of reliable evidence.

\subsection{Evidence Construction Based on Subjective Logic Theory}
\label{Evidence Construction Based on Subjective Logic Theory}
Subjective Logic (SL)~\cite{SL} is a logic-based framework for handling uncertainty and reasoning, originally proposed by Jøsang. It is primarily used for reasoning under conditions of uncertainty, vague decisions, or partial information. In SL, belief vectors are employed to represent both the degree of confidence and uncertainty associated with a given hypothesis. 

As discussed earlier, $m_q(A)$ denotes the credibility of hypothesis $A$ as assessed by channel $q$ of the reference video anchor, acting as an evidence source. We refer to this quantity as the evidence. According to the theory of evidence-based deep learning~\cite{EDL} and SL, the evidence can naturally be represented by the similarity between the semantic vector encoded by channel $q$ of the reference video anchor and the target feature. This is formulated as follows,

\begin{equation}
e_q = m_q(A) = \exp\left( \frac{s_q}{\tau} \right).
\label{evidence_eq}
\end{equation}

In conjunction with the similarity computation, the resulting formulation can be further expressed as follows, which corresponds to \textbf{Equation (8)} in the main paper,

\begin{equation}
e_q = \exp\left( \max_{\hat{q}=1}^{Q} \left( \mathbf{A}_{r(q)} \cdot \mathbf{F}_t^\top \right)_{\hat{q}} \big/ \tau \right),
\label{evidence_max_projection}
\end{equation}
where $\tau$ is the temperature coefficient, $Q$ is the number of learnable queries in the Q-Former, and $F_t$ denotes the target feature. According to SL theory, for each evidence source $m_q$, the corresponding belief vector $E_q$ can be computed as follows,

\begin{equation}
E_q = b_q + u_q,
\label{eq_local_reliability}
\end{equation}
where the belief mass $b_q$ represents the credibility of the hypothesis being valid, and $u_q$ denotes the remaining part of the belief vector. According to SL theory, trust and uncertainty are complementary to each other. Therefore, we have,

\begin{equation}
\sum_{q=1}^{Q} E_q = \sum_{q=1}^{Q} (b_q + u_q) = 1.
\label{eq_belief_uncertainty_total}
\end{equation}

To enhance the credibility of the hypothesis, it is necessary to maximize the total belief mass $b_q$. Based on SL theory, the belief mass $b_q$ can be expressed in terms of the evidence $e_q$ as follows,

\begin{equation}
b_q = \frac{e_q}{S},
\label{belief_from_evidence}
\end{equation}
where $S$ denotes the total strength of evidence, which, according to the Dirichlet distribution, can be represented as the sum of all evidence parameters, that is,

\begin{equation}
S = \sum_{k=1}^{K} \alpha_k = \sum_{k=1}^{K} (e_k + 1).
\label{dirichlet_total_strength}
\end{equation}

By substituting $S$ into the expression for $b_q$, the belief mass $b_q$ can be computed as follows, which corresponds to \textbf{Equation (9)} in the main paper,

\begin{equation}
b_q = \frac{e_q}{\sum_{\hat{q}=1}^{Q} (e_{\hat{q}} + 1)}.
\label{belief_normalized}
\end{equation}

When multiple evidence sources are available, SL allows the fusion of belief vectors using either weighted averaging or Bayesian update. Suppose there are $Q$ evidence sources, each denoted as $m_q$ with an associated belief vector $\mathbf{E}_q$, these belief vectors can be fused as follows,

\begin{equation}
\mathbb{E}_r = \sum_{q=1}^{Q} w_q \, \mathbb{E}_q,
\label{reliability_reference}
\end{equation}
where $w_q$ denotes the weight of each evidence source. In this case, we assume that each channel of the reference video anchor contributes equally to the overall belief mass, that is, $w_q = 1$. By fusing the belief vectors, we obtain the final matching measure between the reference video anchor and the target feature, which we refer to as the correlation reliability, computed as follows,

\begin{equation}
\mathbb{E}_r = \sum_{q=1}^{Q} b_q = 1 - \frac{Q}{\sum_{\hat{q}=1}^{Q} (e_{\hat{q}} + 1)}.
\label{er_closed_form}
\end{equation}

As a result, we obtain the computation of the evidence $\mathbb{E}_r$, which corresponds to \textbf{Equation (10)} in the main paper. According to Equation (7) in the supplementary material, since $e_q \propto S_q$, it follows that $\mathbb{E}_r \propto \sum_{q=1}^{Q} S_q$. Following a similar procedure, the correlation reliability between the modification text anchors and the target feature, denoted as $\mathbb{E}_m$, can also be derived.

Based on the above analysis, we conclude that the correlation reliability values $\mathbb{E}_r$ and $\mathbb{E}_m$ are positively correlated with the similarity between the corresponding anchors and the target feature. Therefore, they can be utilized to optimize the reliability of the similarity between the composed feature and the target feature. Accordingly, the loss function is constructed as follows, which corresponds to \textbf{Equation (11)} in the main paper,

\begin{table*}[ht]
  \centering
  \tabcolsep=2pt

    \begin{tabular}{c|c|c|c|c|c|c}
    \hline
    \hline
    Methods & \multicolumn{1}{c|}{Params (M)} & \multicolumn{1}{c|}{Test (s/sample)} & \multicolumn{1}{c|}{Replication Resource} & \multicolumn{1}{c|}{Train (s/iteration)} & \multicolumn{1}{c|}{WebVid-Avg} & CIRR-Avg \\
    \hline
    CoVR-2 & 1173.19M & 0.0682 & 18541M & 9.423 & 83.30 & 78.92 \\
    ReTrack(w/o\_A ref) & 1176.43M & 0.0682 & 21951M & 9.466 & 84.33 & 79.73 \\
    ReTrack & 1176.43M & 0.0683 & 22885M & 10.946 & 85.70 & 81.09 \\
    \hline 
    \hline
    \end{tabular}%
      \caption{Efficiency Comparison on WebVid-CoVR and CIRR Datasets}
  \label{tab:Efficiency}%
\end{table*}%

\begin{equation}
    \mathcal{L}_{evi}=\frac{1}{B}\sum^B_{b=1}\left(\mathbb{E}_{rb}\!\!-\!\! \mathcal{S}\left(\textbf{F}_{cb}, \textbf{F}_{tb}\right) \right)^2\!\!+\!\!
    \left(\mathbb{E}_{mb}\!\!-\!\!\mathcal{S}\left(\textbf{F}_{cb}, \textbf{F}_{tb}\right) \right)^2,
    \label{belief}
\end{equation}
where $B$ denotes the batch size, and $\textbf{F}_{cb}$ and $\textbf{F}_{tb}$ represent the $b$-th composed feature and target feature in the batch, respectively. In conclusion, we have demonstrated that by computing the reliable evidence between the two sets of directional anchors and the target feature, the reliability of the similarity matrix in the alignment process can be enhanced, thereby reducing the uncertainty in the alignment between the composed feature and the target feature.

\begin{figure}[h]
\begin{center}
\includegraphics[width=0.95\linewidth]{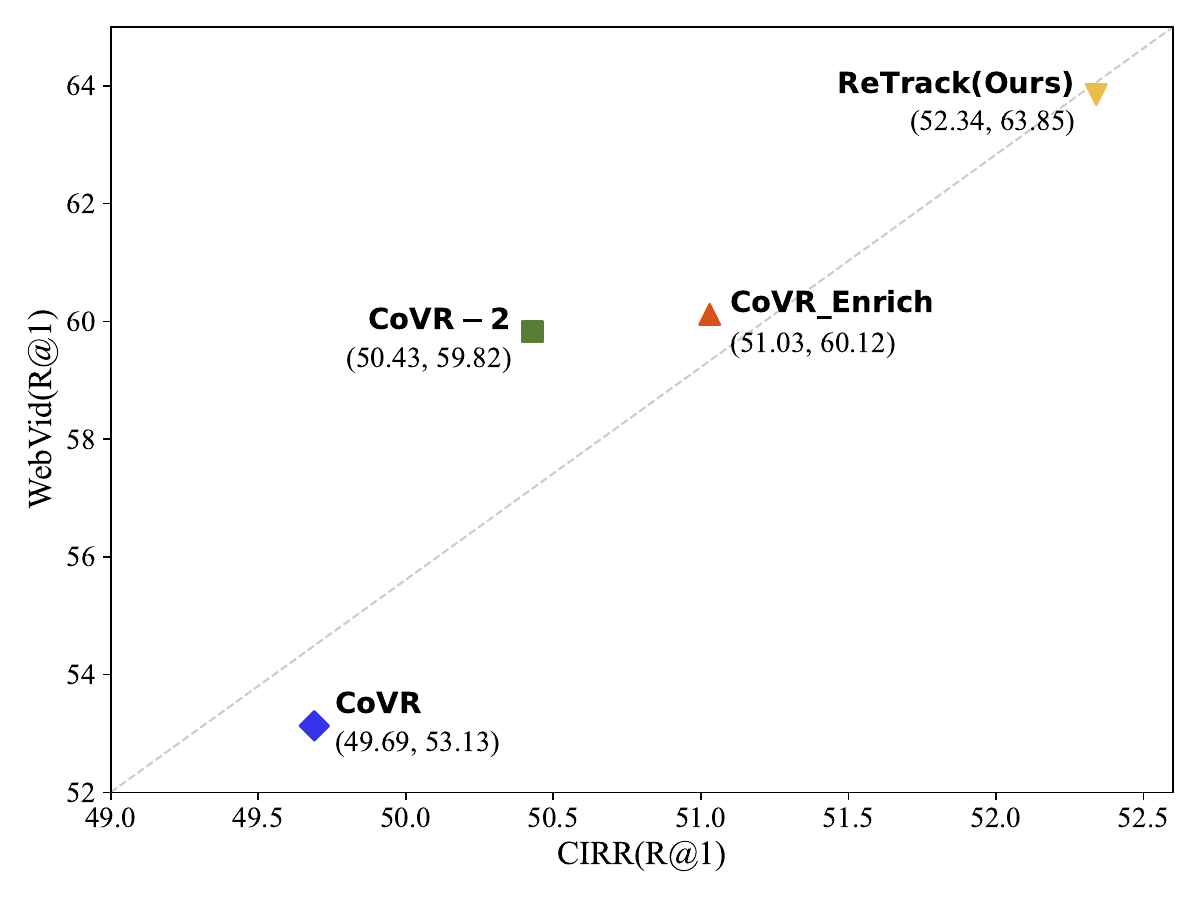}
\end{center}
   \caption{Comprehensive Performance Comparison Between CIR and CVR Tasks}
\label{fig:comprehensive-1}
\end{figure}

\section{Additional Performance Comparison}

\label{appendix:Additional Performance Comparison}

\subsection{Comprehensive Performance Comparison on CIR and CVR Tasks}
\label{appendix:additional_performance_cir_cvr}

In this section, we present a more comprehensive comparison between ReTrack and representative CoVR-2 models, providing an intuitive view of our method's advantages across the CVR and CIR tasks and its potential for practical deployment.

As shown in Figure~\ref{fig:comprehensive-1}, the horizontal axis reports R@1 on the CIRR dataset and the vertical axis reports R@1 on WebVid-CoVR. Compared with models that are applicable to both Composed Video Retrieval (CVR) and Composed Image Retrieval (CIR), including CoVR, CoVR-2, and CoVR\_Enrich, our ReTrack lies in the upper-right region of the plot, indicating leading performance on both CVR and CIR. Furthermore, ReTrack's advantage over CoVR\_Enrich is larger on CIRR than on WebVid-CoVR, which suggests that the proposed \textit{Semantic Contribution Disentanglement} and \textit{Composition Geometry Calibration} (together with \textit{Reliable Evidence-driven Alignment}) provide strong cross-domain generalization by disentangling multimodal semantics and calibrating the composed feature. These results demonstrate ReTrack's superior capability in multimodal semantic understanding.

\subsection{Efficiency Evaluation}
\label{appendix:additional_performance_efficiency}

To comprehensively assess the proposed model, we go beyond conventional recall by conducting efficiency evaluations. Specifically, we examine computational resource consumption during training and inference, processing throughput, and end-to-end response time. These experiments allow us to better gauge real-world performance, particularly under resource-constrained settings.

We therefore compare the efficiency of several models, including our ReTrack, the baseline CoVR-2, and ablated variants of ReTrack. As shown in Table~\ref{tab:Efficiency}, we report parameter counts, training time, and inference latency for the CVR task using the WebVid-CoVR dataset. This comparison makes the efficiency differences among methods explicit and, in conjunction with our method section (i.e., Semantic Contribution Disentanglement, Composition Geometry Calibration, and Reliable Evidence-driven Alignment), clarifies the cost–benefit trade-offs of introducing disentanglement and calibration to achieve robust multimodal retrieval.

\begin{algorithm}[h!]
\small
\caption{ReTrack Training Procedure}
\label{alg:retrack}
\begin{algorithmic}[1]
\Require Triplets $\mathcal{T}=\{(x_r,x_m,x_t)_n\}_{n=1}^{N}$; encoders $\varPhi_{\mathbb{I}},\varPhi_{\mathbb{T}}$; Q-Former; Transformer-Decoder; MLP; similarity $\mathcal{S}(\cdot,\cdot)$; temperature $\tau$; weights $\kappa,\lambda$; batch size $B$; optimizer (e.g., Adam); max epochs $E$
\Ensure Trained parameters $\Theta^{*}$ of ReTrack
\State Initialize $\Theta$
\For{$e=1$ to $E$}
  \State Shuffle $\mathcal{T}$
  \For{\textbf{each} minibatch $\{(x_r^{i},x_m^{i},x_t^{i})\}_{i=1}^{B}$}
    \State \textbf{Bimodal Extraction \& Composition (Sec.~3.2)}$\rightarrow$
    \State Obtain $\mathbf{F}_r^{i},\mathbf{F}_m^{i}, \mathbf{F}_c^{i}, \mathbf{F}_t^{i}$
    \State \textbf{Contribution Disentanglement (Sec.~3.2)}$\rightarrow$
    \State $\mathbf{P}_r^{i}=\operatorname{Decoder}(Q=\mathbf{F}_r^{i},\{K,V\}=\mathbf{F}_c^{i})$
    \State $\mathbf{P}_m^{i}=\operatorname{Decoder}(Q=\mathbf{F}_m^{i},\{K,V\}=\mathbf{F}_c^{i})$
    \State \textbf{Anchor Generation (Sec.~3.3)}$\rightarrow$
    \State $\mathbf{W}_{p}^{r,i}\!\!=\!\!\operatorname{MLP}(\mathbf{F}_c^{i}(\mathbf{F}_r^{i})^{\top}),
           \mathbf{W}_{p}^{m,i}\!\!=\!\!\operatorname{MLP}(\mathbf{F}_c^{i}(\mathbf{F}_m^{i})^{\top})$
    \State $\mathbf{A}_r^{i}\!\!=\!\!\mathbf{F}_c^{i}+\mathbf{W}_{p}^{r,i}\odot \mathbf{P}_r^{i},
    \mathbf{A}_m^{i}\!\!=\!\!\mathbf{F}_c^{i}+\mathbf{W}_{p}^{m,i}\odot \mathbf{P}_m^{i}$
    \State \textbf{Distance-oriented Alignment (Sec.~3.3)}$\rightarrow$
    \State $\displaystyle
      \mathcal{L}_{dis}=\frac{1}{B}\sum_{i=1}^{B} -\log \frac{\exp\{\mathcal{S}(\mathbf{F}_{c}^{i},\mathbf{F}_{t}^{i})/\tau\}}
             {\sum_{j=1}^{B}\exp\{\mathcal{S}(\mathbf{F}_{c}^{i},\mathbf{F}_{t}^{j})/\tau\}}$
    \State \textbf{Direction-oriented Calibration (Sec.~3.3)}
    \State $\mathbf{A}_{c}^{i}=(\mathbf{A}_{r}^{i}-\mathbf{F}_{c}^{i})+(\mathbf{A}_{m}^{i}-\mathbf{F}_{c}^{i}),\quad
           \mathbf{A}_{t}^{i}=\mathbf{F}_{t}^{i}-\mathbf{F}_{c}^{i}$
    \State $\displaystyle
      \mathcal{L}_{dir}=\frac{1}{B}\sum_{i=1}^{B} -\log \frac{\exp\{\mathcal{S}(\mathbf{A}_{c}^{i},\mathbf{A}_{t}^{i})/\tau\}}
             {\sum_{j=1}^{B}\exp\{\mathcal{S}(\mathbf{A}_{c}^{i},\mathbf{A}_{t}^{j})/\tau\}}$
    \State \textbf{Reliable Evidence-driven Alignment (DST/EDL)}$\rightarrow$
    \For{$i=1$ to $B$} \Comment{Evidence modeling per sample}
      \For{$q=1$ to $Q$}
        \State $e_{r,q}^{i}=\exp\Big(\max_{\hat q}\big(\mathbf{A}_{r(q)}^{i}(\mathbf{F}_{t}^{i})^{\top}\big)_{\hat q}/\tau\Big)$
        \State $e_{m,q}^{i}=\exp\Big(\max_{\hat q}\big(\mathbf{A}_{m(q)}^{i}(\mathbf{F}_{t}^{i})^{\top}\big)_{\hat q}/\tau\Big)$
        \State $b_{r,q}^{i}\!\!=\!\!\dfrac{e_{r,q}^{i}}{\sum_{\hat q=1}^{Q}\!(e_{r,\hat q}^{i}\!+\!1)},
               b_{m,q}^{i}\!\!=\!\!\dfrac{e_{m,q}^{i}}{\sum_{\hat q=1}^{Q}\!(e_{m,\hat q}^{i}\!+\!1)}$
      \EndFor
      \State $\mathbb{E}_{r}^{i}=\sum_{q=1}^{Q} b_{r,q}^{i},\quad \mathbb{E}_{m}^{i}=\sum_{q=1}^{Q} b_{m,q}^{i}$
    \EndFor
    \State \textbf{Evidence regularization (Sec.~3.4)}$\rightarrow$
    \State $\displaystyle
      \mathcal{L}_{evi}\!\!=\!\!\frac{1}{B}\sum_{i=1}^{B}\Big(\mathbb{E}_{r}^{i}\!\!-\!\!\mathcal{S}(\mathbf{F}_{c}^{i},\mathbf{F}_{t}^{i})\Big)^{2}
      +\Big(\mathbb{E}_{m}^{i}-\mathcal{S}(\mathbf{F}_{c}^{i},\mathbf{F}_{t}^{i})\Big)^{2}$
    \State \textbf{Overall objective \& update (Eq.(12)))}$\rightarrow$
    \State $\mathcal{L}=\mathcal{L}_{dis}+\kappa\,\mathcal{L}_{dir}+\lambda\,\mathcal{L}_{evi}$
    \State $\Theta \leftarrow \operatorname{OptimizerUpdate}(\Theta,\nabla_{\Theta}\mathcal{L})$
  \EndFor
\EndFor
\State \Return $\Theta^{*}$
\end{algorithmic}
\end{algorithm}

\begin{figure*}[h!]
\begin{center}
\includegraphics[width=0.8\linewidth]{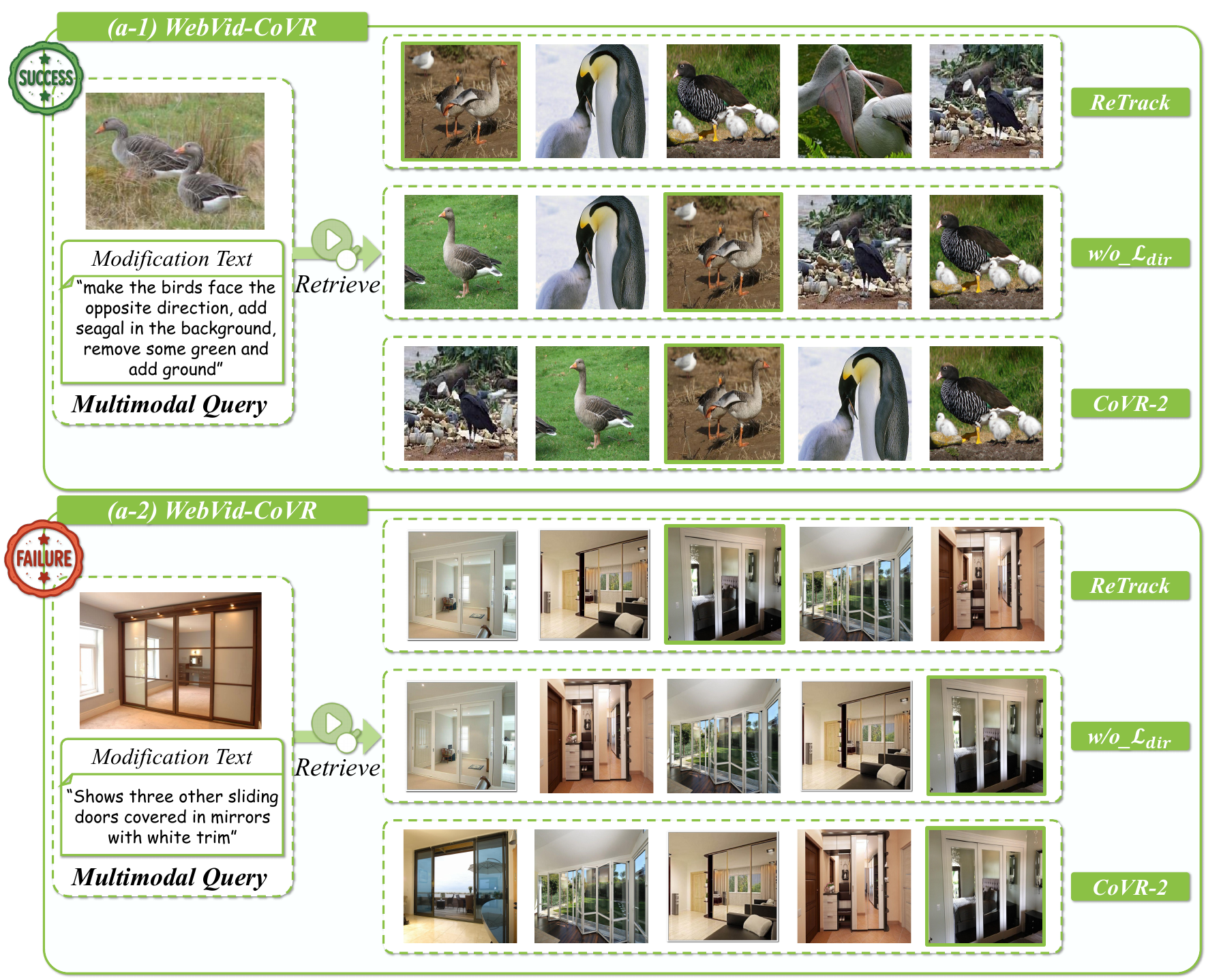}
\end{center}
   \caption{More Cases on CVR task.}
\label{fig:case_CVR}
\end{figure*}

\begin{figure*}[h!]
\begin{center}
\includegraphics[width=0.7\linewidth]{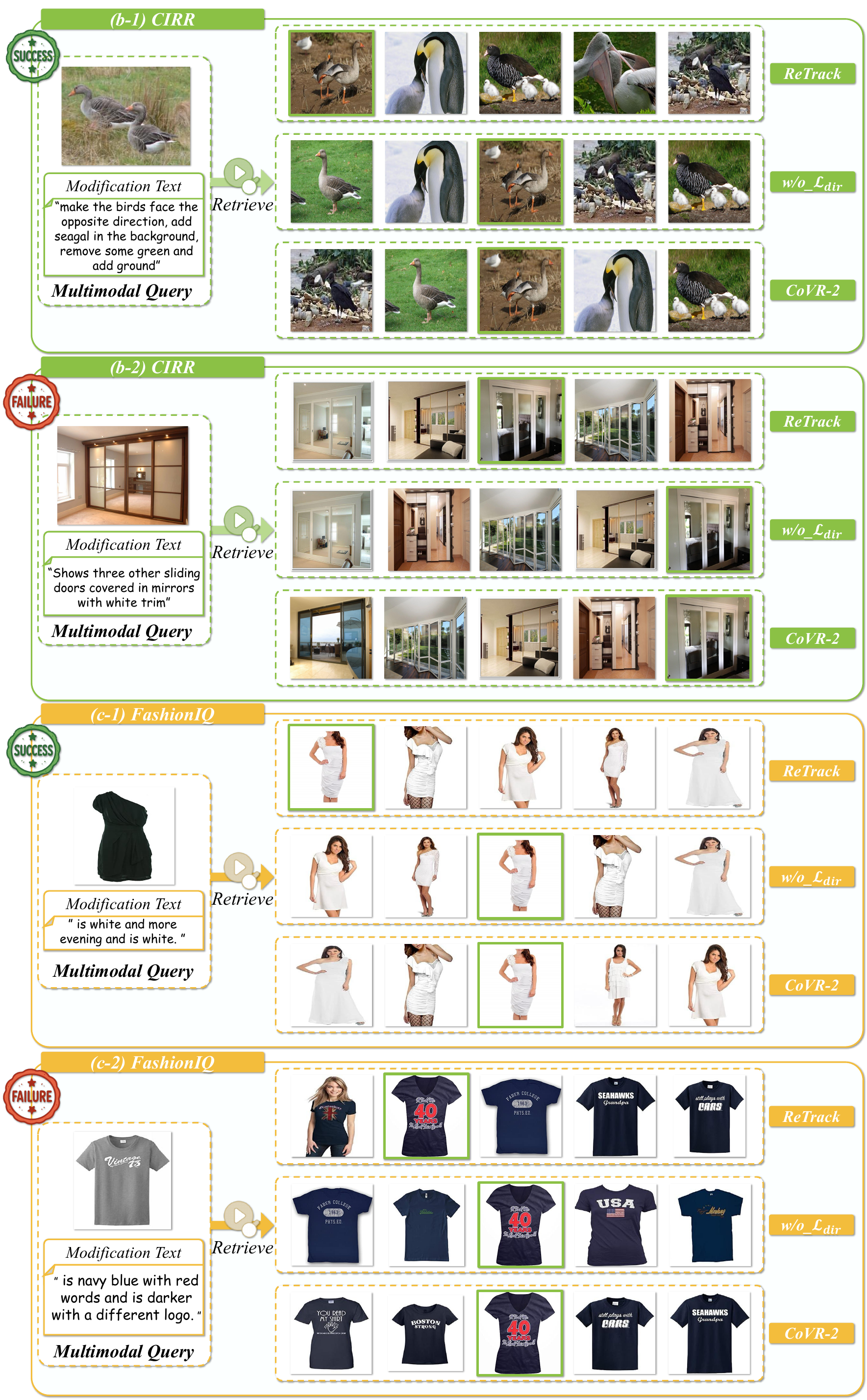}
\end{center}
   \caption{More Cases on CIR task.}
\label{fig:case_CIR}
\end{figure*}

In terms of parameter count, ReTrack and its variant without the reference anchor (w/o A\_ref) add approximately 3M parameters over CoVR-2; nevertheless, this increase does not introduce a noticeable performance penalty, indicating that ReTrack remains relatively compact while delivering superior retrieval accuracy. For inference, the three methods exhibit nearly identical latency (~$0.068$ s), showing that ReTrack and its variant match CoVR-2's runtime efficiency despite architectural changes. Training time is moderately higher for ReTrack, and some variants such as w/o A\_ref are almost on par with CoVR-2, whereas the standard model incurs about +1.5s per iteration. This overhead is expected and reasonable because ReTrack introduces Semantic Contribution Disentanglement, Composition Geometry Calibration (with dual directional anchors and a direction-aware loss), and Reliable Evidence-driven Alignment, which together add calibration and regularization steps to better model the composed feature. Regarding compute resources, ReTrack and its variants consume more than CoVR-2 due to processing richer multimodal semantics (via disentanglement), constructing and calibrating directional anchors, and applying evidence-driven regularization. However, the resulting accuracy gains outweigh these costs. Taken together across parameter count, inference latency, training time, and resource usage, ReTrack achieves a favorable cost–performance trade-off, maintaining CoVR-2–level efficiency at inference while providing markedly stronger retrieval performance.

\section{Algorithm of Retrack's Training Procedure}
\label{appendix:algorithm}

To complement the main methodology, we provide the full training procedure of ReTrack in the form of pseudocode below. This offers a clear and reproducible description of how the disentanglement, calibration, and evidence-driven alignment modules are jointly optimized during training.

\section{More Case Study}
\label{appendix:case}

In the supplementary material, to further validate the effectiveness of ReTrack and the contribution of its key loss function, Direction-oriented Calibration ($\mathcal{L}_{dir}$), we present qualitative retrieval results on three datasets: WebVid-CoVR (for the CVR task), and CIRR and FashionIQ (for the CIR task). For each dataset, we include one successful case and one failure case from three different models: the complete ReTrack model, a variant without the direction calibration loss $\mathcal{L}_{dir}$ (denoted as w/o $\mathcal{L}_{dir}$), and the representative baseline method CoVR-2.

\textbf{(a-1) WebVid-CoVR dataset \underline{Successful} Case:} For the modification text “change the season to springtime,” only ReTrack successfully retrieves scenes characteristic of spring (e.g., budding trees and green grass), accurately capturing the seasonal semantic transition. This effectiveness is attributed to our proposed Composition Geometry Calibration module, which aligns the semantic shift from “autumn” to “spring” through the construction of directional anchors. In contrast, other models either return seasonally inconsistent scenes or produce results dominated by repeated visual elements (e.g., trees) from the reference video, leading to semantic distortion.  

\textbf{(a-2) WebVid-CoVR dataset \underline{Failure} Case:} For the modification “change it to cappuccino,” none of the models successfully retrieve the correct result among their top-ranked outputs. This may be due to inconsistent annotations for beverage categories or visual ambiguity in the dataset (e.g., variations in angle or cup shape), rather than model deficiency. Notably, ReTrack and its variant still retrieve scenes containing milk or coffee-related content, suggesting a certain level of directional understanding.

\textbf{(b-1) CIRR dataset \underline{Successful} Case:} For the modification “make the birds face the opposite direction, add seagull in the background, remove some green and add ground,” ReTrack successfully adjusts both orientation and background elements, demonstrating its capacity to model complex spatial transformations in natural scenes. This is attributed to the direction-oriented calibration mechanism, which refines the compositional semantic direction, enabling the composed feature to better align with the spatial structure of the target image while still accurately capturing multiple textual modification cues. In contrast, other models retrieve visually similar but directionally incorrect images.  

\textbf{(b-2) CIRR dataset \underline{Failure} Case:} The textual description “Shows three other sliding doors...” is highly specific, yet this particular viewpoint or instance may be absent from the CIRR dataset. Consequently, none of the models successfully retrieve a correct match. This limitation is more likely due to insufficient sample coverage or incomplete annotations in the dataset rather than deficiencies in the models’ reasoning capabilities.

\textbf{(c-1) FashionIQ dataset \underline{Successful} Case:} Given the modification “is white and more evening and is white,” ReTrack retrieves multiple white evening dresses that align well with the described style. This performance results from the effective modeling of the Semantic Contribution Disentanglement module, which accurately extracts the modification semantics from the text and integrates them into the composed feature through directional reconstruction.

\textbf{(c-2) FashionIQ dataset \underline{Failure} Case:} The user query “is navy blue with red words and is darker with a different logo” involves changes to textual patterns and color schemes. However, the FashionIQ dataset lacks detailed annotations for textual graphics or high-resolution logos, leading to failure across all models in recognizing such differences. Nevertheless, ReTrack retrieves results that are closer to the target in terms of color and style, demonstrating partial capability in addressing this type of query.

Across the three datasets, ReTrack consistently outperforms both the ablated variant and the baseline method in retrieval tasks involving complex semantics or directional changes. Most of the failures can be attributed to missing annotations, ambiguous query descriptions, or limited sample coverage, rather than deficiencies in the model design itself. These case studies further validate the necessity and robustness of the Direction-oriented Calibration and Evidence-driven Alignment modules in ReTrack, particularly highlighting their advantages in addressing challenges such as modality bias and semantic ambiguity.

\small
\bibliography{aaai2026}

\end{document}